\documentclass[sigconf,nonacm]{acmart}
\setcopyright{none}
\acmConference[WSDM '27]{The 20th ACM International Conference on Web Search and Data Mining}
  {February 15--19, 2027}{Hong Kong, China}
\acmYear{2027}
\renewcommand\footnotetextcopyrightpermission[1]{}
\usepackage{listings}
\usepackage[normalem]{ulem}
\usepackage{xcolor}
\usepackage{multirow}
\usepackage{amsmath}
\usepackage{booktabs}
\usepackage{algorithm}
\usepackage{colortbl}
\usepackage[skins,breakable]{tcolorbox}

\newcommand{\method}{CoMem}

\title{Beyond Compressing Everything: Contrastive Memory Learning for Selective Context Compression}

\author{Sikui Zhang}
\affiliation{\institution{Institute of Automation, Chinese
Academy of Sciences, China}\country{}}
\author{Haowei Liu}
\affiliation{\institution{Alibaba Group}\country{China}}
\author{Ziyun Gan}
\affiliation{\institution{Institute of Automation, Chinese
Academy of Sciences, China}\country{}}
\author{Chunfeng Yuan}
\authornote{Corresponding author.}
\affiliation{\institution{Institute of Automation, Chinese
Academy of Sciences, China}\country{}}
\author{Guangze Gao}
\affiliation{\institution{Institute of Automation, Chinese
Academy of Sciences, China}\country{}}
\author{Nian Liu}
\affiliation{\institution{Institute of Automation, Chinese
Academy of Sciences, China}\country{}}
\author{Boyu Cai}
\affiliation{\institution{Institute of Automation, Chinese
Academy of Sciences, China}\country{}}
\author{Bo Li}
\affiliation{\institution{Hello Group}\country{China}}
\author{Xiaolei Lv}
\affiliation{\institution{Hello Group}\country{China}}
\author{Jun Gao}
\affiliation{\institution{Hello Group}\country{China}}
\author{Bing Li}
\affiliation{\institution{Institute of Automation, Chinese
Academy of Sciences, China}\country{}}
\author{Weiming Hu}
\affiliation{\institution{Institute of Automation, Chinese
Academy of Sciences, China}\country{}}
\renewcommand{\shortauthors}{Zhang et al.}
\begin{document}

\begin{abstract}
As large language models (LLMs) advance toward increasingly long inputs, soft context compression offers a promising method to efficient inference by converting long inputs into compact memory slots. Under limited memory budgets, effective context compression remains challenging because task-relevant evidence  varies substantially across queries,   ranging from localized evidence to multiple distant spans or information dispersed throughout the full document. Instead of compressing everything indiscriminately, we propose \method{}, a query-guided selective context compression framework that learns memory representations conditioned on both the document and  query.  \method{} develops Contrastive Compressor Learning to achieve query-sensitive yet context-preserving memory representations. To inject a strong prior over query-dependent evidence distributions, we  devise a same-context hard negative mining strategy by pairing a single document with multiple queries, requiring the compressor to selectively retain information according to each query’s distinct evidence needs.  To address the restricted information flow imposed by causal masking, we further enable bidirectional attention among memory slots. Experiments demonstrate that \method{} consistently outperforms existing context compression methods across compression ratios on question answering and RULER-NIAH benchmarks. Our code is available at \href{https://github.com/scar-on/CoMem}{\textcolor{blue}{https://github.com/scar-on/CoMem}}.
    \end{abstract}

\keywords{Context Compression, Large Language Models, Question Answering}

\maketitle

\definecolor{stringBlue}{RGB}{51,202,246}

\section{Introduction}

\begin{figure}[!tb]
    \centering
    \includegraphics[width=0.99\linewidth]{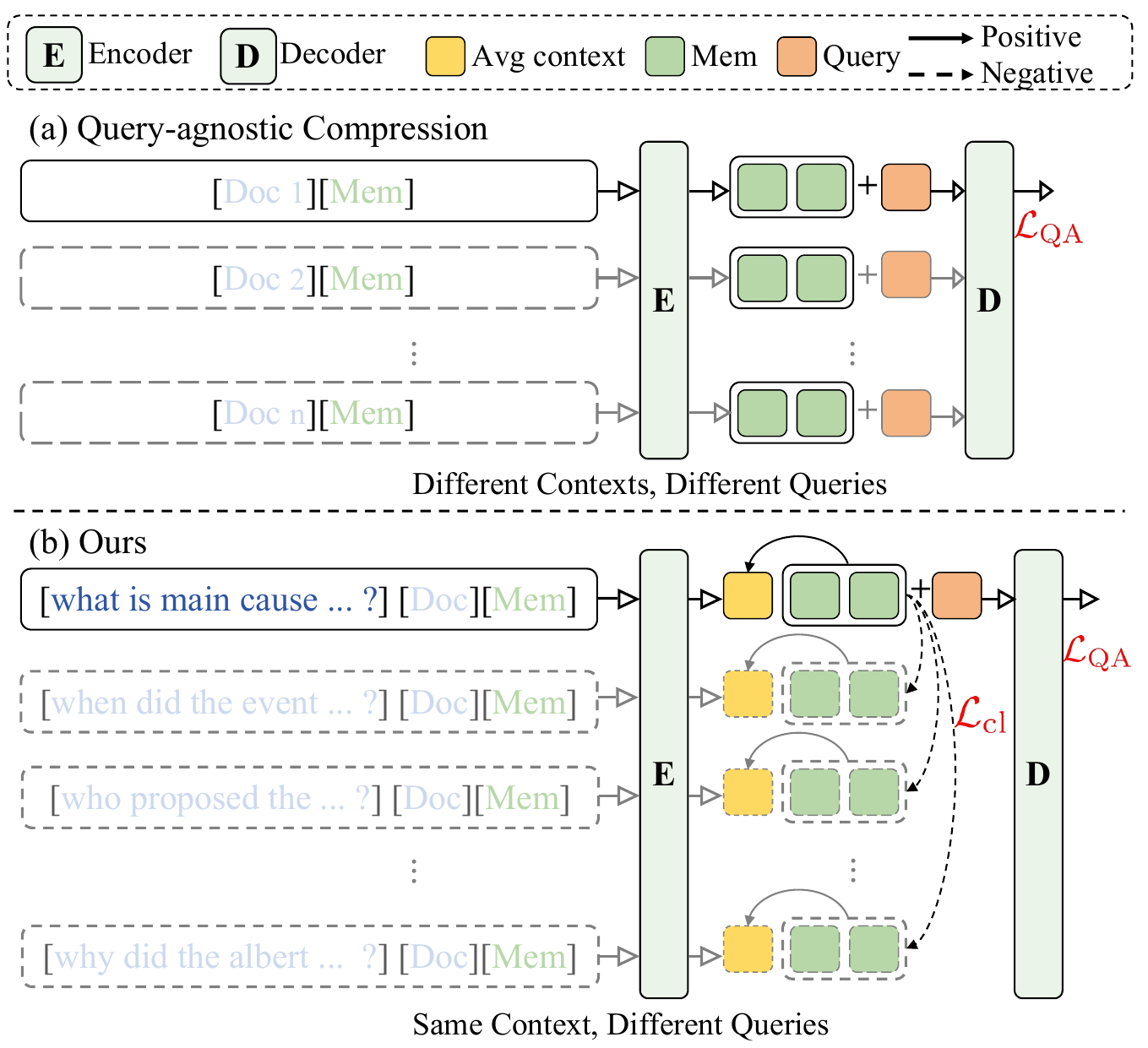}
    \caption{Comparison between the query-agnostic compression and \method{}.
    (a) The query-agnostic method  learns memory slots supervised
     by the QA loss. (b) Our method further introduces 
    contrastive compressor learning and trains on different queries from the same context.}
    \Description{Comparison between the query-agnostic compression and \method{}.
    (a) The query-agnostic method  learns memory slots supervised
     by the QA loss. (b) Our method further introduces 
    contrastive compressor learning and trains on different queries from the same context.}
    \label{fig:method_compare}
    \vspace{-2pt}
\end{figure}

\begin{figure*}[!tb]
    \centering
    \includegraphics[width=0.99\linewidth]{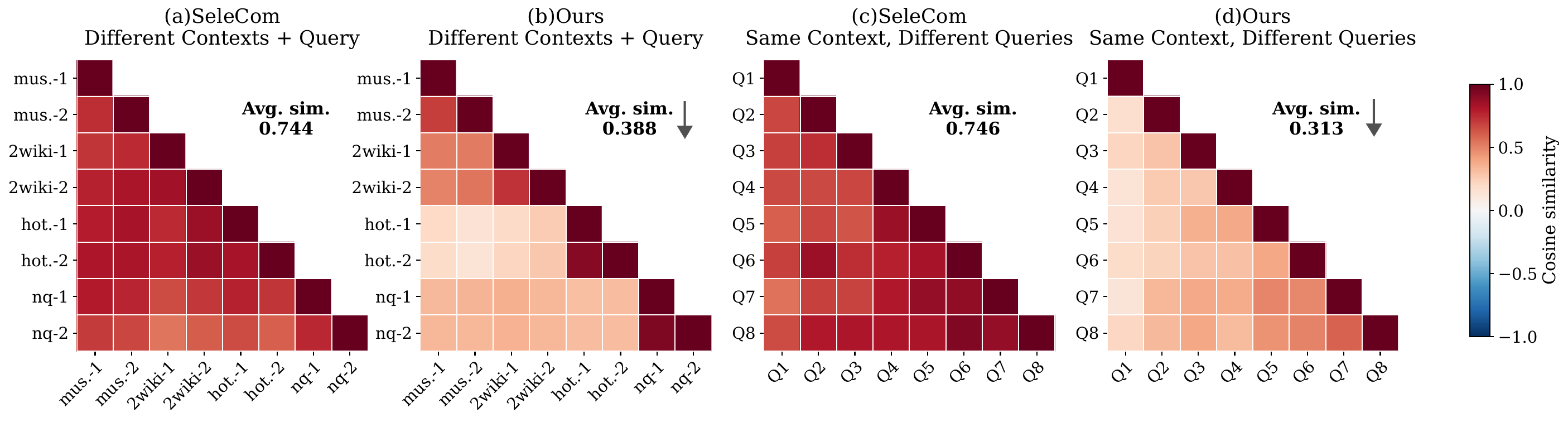}
    \caption{
    Similarity patterns of compressed memory slots.
  (a) and (b) show memory slots from different contexts, (c) and (d) show  memory slots from the
same context under different queries. In both settings, our method achieves lower inter-slot similarity,
demonstrating more discriminative query-guided compression.
    }
    \Description{
    Similarity patterns of compressed memory slots.
  (a) and (b) show memory slots from different contexts, (c) and (d) show  memory slots from the
same context under different queries. In both settings, our method achieves lower inter-slot similarity,
demonstrating more discriminative query-guided compression.
    }
    \label{fig:four_heatmaps}
\end{figure*}

Large language models (LLMs) have substantially expanded the
amount of information that can be provided to a model at inference time.
This is particularly important for long-document question answering~\cite{li2024long,wang2025novelqa} and
retrieval-augmented generation (RAG)~\cite{gao2025d}, where the answer often depends on
evidence distributed across a large context. Yet long context does not mean
effective information use, natural language is inherently redundant~\cite{wit1999linguistic,li2023compressing}.  Taking the entire document as input to an LLM may introduce inference cost, distractors~\cite{liu2024lost}, dilute attention over irrelevant content, and substantially increase computation and KV-cache overhead~\cite{NEURIPS2024_5dfbe6f5,han2024lm}. Soft context
compression~\cite{li2025prompt} offers a promising alternative by compressing full context into 
short continuous memory slots, enabling the decoder to reason
over a compact representation rather than the full document.

Existing soft compression methods, such as ICAE~\cite{ge2024incontext}, 500x~\cite{li2025500xcompressor}, PCC~\cite{dai2025pretraining}, and SAC~\cite{liu2026iclr-autoencodingfree}, typically compress a document into a fixed memory representation independent of the downstream query. However, the essence of compression is not merely to reduce the number of tokens, but to determine what information should be preserved under a limited memory budget. The same context may contain critical evidence for one question while serving only as background for another. Recent works have begun to concatenate the query with the document  for compression~\cite{louis2026oscar,Liu2026SeleCom}. Although the query is incorporated into the compression input, this design does not explicitly encourage memory slots to specialize according to query needs. Consequently, in practice, the learned memory slots remain relatively query-agnostic, leaving most query-sensitive reasoning to the decoder. This limitation is illustrated in Figure~\ref{fig:four_heatmaps}(a, c), where the SeleCom~\cite{Liu2026SeleCom} compressor generates highly similar memory slots not only across different contexts but also across different queries paired with the same context. These findings highlight the need for stronger mechanisms to explicitly encourage query-guided memory specialization.

In this paper, we argue that effective query-guided context compression requires stronger supervision than simple query concatenation. As illustrated in Figure~\ref{fig:method_compare}, compressed memory representations should remain distinguishable under different queries even for the same context. To this end, we present \method{}, a novel architecture for query-guided selective context compression. \method{} develops Contrastive Compressor Learning to preserve contextual information by aligning memory slots with their corresponding context representations, while separating representations from different contexts. To further strengthen query-guided compression signals, we devise a same-context hard negative mining strategy that contrasts memory slots from different queries over the same context, encouraging query-sensitive yet context-preserving memory representations. To enable richer information flow and more effective information allocation during compression, we allow memory slots to interact bidirectionally. Concretely, we introduce bidirectional attention among memory slots while preserving causal masking for the original query-context sequence. By combining Contrastive Compressor Learning with Bidirectional Flow Memory, \method{} enables more effective query-guided compression. We conduct a comprehensive evaluation of \method{} on four question-answering benchmarks and RULER-NIAH. \method{} achieves superior overall performance across both QA and retrieval tasks. Notably, under 4$\times$ compression, it even surpasses the original context in LLM-as-a-judge scores on several benchmarks, while remaining competitive even at 64$\times$ compression.

 \begin{figure*}[!tb]
    \centering
    \includegraphics[width=0.95\linewidth]{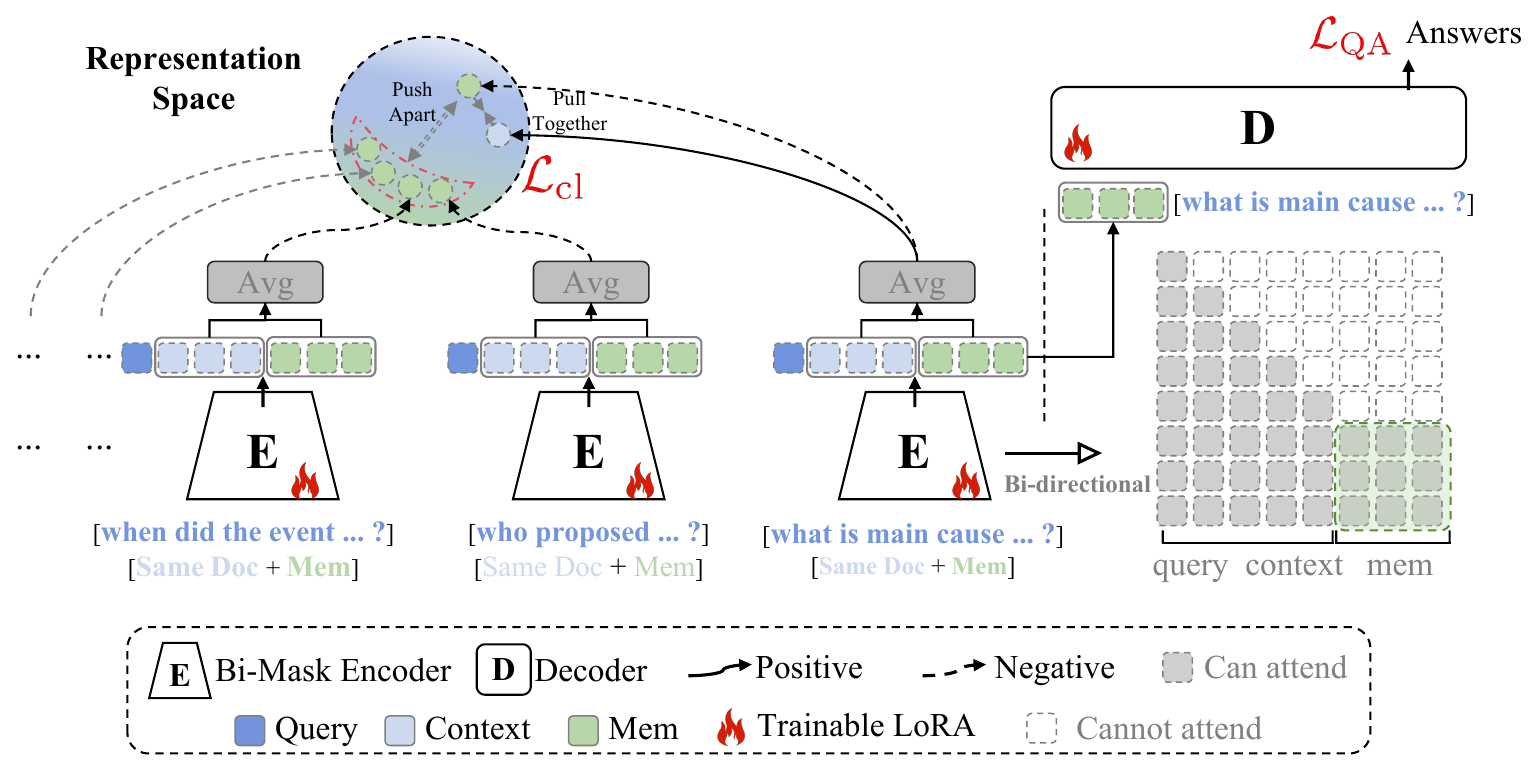}
    \caption{Overview of \method{}.  The QA objective \textcolor{red}{$\mathcal{L}_{\mathrm{QA}}$} provides
    an end-task generation signal, while the contrastive
    learning objective \textcolor{red}{$\mathcal{L}_{\mathrm{cl}}$}
    imposes a more explicit representation-level signal, aligning memory
    slots with their source context while using same-context,
    different-query examples as hard negatives to encourage query-guided
    memory representations.}
    \Description{Overview of \method{}.  The QA objective \textcolor{red}{$\mathcal{L}_{\mathrm{QA}}$} provides
    an end-task generation signal, while the contrastive
    learning objective \textcolor{red}{$\mathcal{L}_{\mathrm{cl}}$}
    imposes a more explicit representation-level signal, aligning memory
    slots with their source context while using same-context,
    different-query examples as hard negatives to encourage query-guided
    memory representations.}
    \label{fig:main}
\end{figure*}

Our contributions are summarized as follows:
\begin{itemize}

    \item We propose \method{}, a  query-guided selective context compression framework that learns memory slots that are consistent with their source context
and distinct across different queries over the same context.

    \item We develop contrastive compressor learning with same-context hard negatives to encourage query-sensitive compression, together with a bidirectional memory interaction mechanism for more effective information allocation.

    \item Extensive experiments on four question-answering benchmarks and RULER-NIAH show that \method{} consistently outperforms prior methods and remains effective under high compression ratios.

\end{itemize}

\section{Method}
\subsection{Task Formulation}
Given a long document context $C = (c_1,\ldots,c_n)$, a query
$Q = (q_1,\ldots,q_l)$, and a target answer
$Y = (y_1,\ldots,y_T)$, our goal is to generate the answer without
feeding the full context into the decoder. Instead, we learn a
query-conditioned compressor that maps the query-context pair into a
compact set of memory representations.
We construct the encoder input in Eq.~\ref{eq:encoder-input} by appending
$K$ learnable memory tokens to the query-context sequence:
\begin{equation}
\label{eq:encoder-input}
X_{\mathrm{enc}} =
[\texttt{<ENC>}, Q, C,
\texttt{<MEM}_1\texttt{>}, \ldots, \texttt{<MEM}_K\texttt{>}].
\end{equation}
The encoder processes $X_{\mathrm{enc}}$ and returns a hidden state for each
token at every layer. We use the final-layer hidden states corresponding to the
last $K$ memory tokens as the compressed representation in
Eq.~\ref{eq:memory-representation}:
\begin{equation}
\label{eq:memory-representation}
M = f_{\phi}(C, Q)
= [h^{m}_1,\ldots,h^{m}_K]
\in \mathbb{R}^{K \times d},
\end{equation}
where $h_k^m \in \mathbb{R}^{d}$ is the final-layer hidden state of the
$k$-th memory token and $K \ll n$. Thus, $M$ acts as a compact information
bottleneck over the original context. Unlike query-agnostic compression, the memory
representation is conditioned on both the document and the query, so the
compressor is encouraged to preserve the context information most relevant
to answering the given query.
The decoder then generates the answer from the query and the compressed
memory representation as shown in Eq.~\ref{eq:compressed-decoding}:
\begin{equation}
\label{eq:compressed-decoding}
p_{\theta}(Y \mid C, Q)
\approx
p_{\theta}(Y \mid Q, M),
\end{equation}
where the conditional distribution is factorized in
Eq.~\ref{eq:decoder-factorization}:
\begin{equation}
\label{eq:decoder-factorization}
p_{\theta}(Y \mid Q, M)
=
\prod_{t=1}^{T}
p_{\theta}(y_t \mid y_{<t}, Q, M).
\end{equation}
At inference time, the model follows a two-stage procedure: it first
compresses the query-context pair into memory slots $M$, and then
conditions the decoder on $Q$ and $M$ to generate the final answer. 

\subsection{Contrastive Compressor Learning}
The supervised generation objective only indirectly supervises the compressor.
As long as the decoder can recover the answer from the memory slots, the encoder
is not explicitly encouraged to form faithful or discriminative memory
representations. This can lead to overly similar memories across inputs with
shared linguistic patterns or answer formats.  To address this limitation, we develop Contrastive Compressor Learning,
as shown in Figure~\ref{fig:main}, with two components: (1) memory-context alignment for preserving contextual information in compressed memory slots and
separating memory representations from different contexts; and (2) same-context hard negative
mining for injecting a strong prior over query-dependent evidence distributions
by contrasting memories from the same context under different queries.

\subsubsection{Memory-Context Alignment.}
To instantiate this objective, for the $i$-th training instance
$(C_i, Q_i, Y_i)$, the encoder produces $K$ final-layer memory hidden states
$[h^m_{i,1}, \ldots, h^m_{i,K}]$, consistent with
Eq.~\ref{eq:memory-representation}:
\begin{equation}
\label{eq:instance-memory}
M_i = f_{\phi}(C_i, Q_i)
=
[h^m_{i,1}, \ldots, h^m_{i,K}]
\in \mathbb{R}^{K \times d}.
\end{equation}

We then mean pool these memory states into a single compressed
representation, as shown in Eq.~\ref{eq:memory-pooling}:
\begin{equation}
\label{eq:memory-pooling}
\mathbf{m}_i = \mathrm{MeanPool}(M_i) \in \mathbb{R}^{d}.
\end{equation}
In parallel, we collect the final-layer encoder hidden states
$H_i^{c}=[h_{i,1}^{c},\ldots,h_{i,n}^{c}]$ corresponding to
the context tokens and mean pool them in Eq.~\ref{eq:context-pooling}:
\begin{equation}
\label{eq:context-pooling}
\mathbf{c}_i = \mathrm{MeanPool}(H^{c}_i) \in \mathbb{R}^{d}.
\end{equation}

As illustrated in Figure~\ref{fig:main}, in the representation space, the positive pair is formed between the compressed memory representation
and the context representation from the same instance,
$(\mathbf{m}_i, \mathbf{c}_i)$. Pulling this pair together encourages the
memory bottleneck to preserve the semantic information of its source
context. In contrast, the negative pairs are formed among different memory
representations in the mini-batch, i.e.,
$(\mathbf{m}_i, \mathbf{m}_j)$ for $j \neq i$. Pushing these memory
representations apart prevents the compressor from mapping different
query-context pairs into indistinguishable memory states.
Formally, the contrastive objective is defined in
Eq.~\ref{eq:contrastive-loss}:
\begin{equation}
\label{eq:contrastive-loss}
\begin{aligned}
\mathcal{L}_{\mathrm{cl}}
&=
-\frac{1}{N}
\sum_{i=1}^{N}
\log
\frac{
\exp(s(\mathbf{m}_i,\mathbf{c}_i)/\tau)
}{
\sum_{\mathbf{z} \in \mathcal{Z}_i}
\exp(s(\mathbf{m}_i,\mathbf{z})/\tau)
},
\\
\mathcal{Z}_i
&=
\{\mathbf{c}_i\}\cup\{\mathbf{m}_j:j\neq i\}.
\end{aligned}
\end{equation}
Here, $s(\cdot,\cdot)$ denotes cosine similarity and $\tau$ is the
temperature, and $N$ is the mini-batch size. The set $\mathcal{Z}_i$ contains one positive context
representation and all negative memory representations in the mini-batch.
Thus, unlike conventional contrastive learning that contrasts an anchor
against mismatched context representations, our objective uses the
memory slots $\mathbf{m}_i$ as the contrastive query. The
corresponding context representation $\mathbf{c}_i$ serves as a
stop-gradient positive target, and the other compressed memories serve as
negatives. Since gradients from $\mathcal{L}_{\mathrm{cl}}$ do not update
$\mathbf{c}_i$, the loss directly optimizes memory representations to
align with their source contexts while occupying distinguishable regions
of the representation space.

\subsubsection{Same-Context Hard Negative Mining Strategy.}
A naive in-batch contrastive objective often uses examples from different
contexts as negatives. However, different contexts are usually already
well-separated in representation space, making such negatives relatively
easy. As a result, the contrastive objective can converge quickly without
forcing the compressor to learn the more important behavior of selective
compression. This tendency toward rapid convergence is illustrated in Figure~\ref{fig:loss_compare}.

To address this, we construct each mini-batch using the same context paired
with different queries. Thus, all negative memories in the batch come from
the same document but are conditioned on different queries. For an anchor
memory $\mathbf{m}_i$, its positive is the context representation
$\mathbf{c}_i$, while its negatives are the other memory representations
$\{\mathbf{m}_j\}_{j \neq i}$ from the same context.
This design turns query variation into hard negatives. Since all memories
come from the same context, the model cannot rely on coarse document-level
differences to separate them. Instead, it must encode the query-specific
information needed to distinguish which aspects of the shared context are
compressed into each memory.

\subsubsection{Overall Training Objective.} We train the model jointly with the supervised answer generation objective.
Given the  memory slots $M_i=f_{\phi}(C_i,Q_i)$, the QA loss is the
negative log-likelihood of the target answer, defined in
Eq.~\ref{eq:qa-objective}:
\begin{equation}
\label{eq:qa-objective}
\mathcal{L}_{\mathrm{QA}}
=
-\sum_{t=1}^{T}
\log p_{\theta}
\left(
y_t \mid y_{<t}, Q_i, f_{\phi}(C_i,Q_i)
\right).
\end{equation}
The final training objective combines answer supervision with the
same-context contrastive objective:
\begin{equation}
\label{eq:total-objective}
\mathcal{L}
=
\mathcal{L}_{\mathrm{QA}}
+
\lambda \mathcal{L}_{\mathrm{cl}},
\end{equation}
where $\lambda$ controls the strength of the contrastive compressor
learning signal.

\subsection{Bidirectional Flow Memory}

Prior work shows that attention masks strongly shape information flow~\cite{wei2026deepseek,behnamghader2024llmvec}. DeepSeek OCR2
shows that tailored attention masks can help learnable output tokens distill and
reorganize source information, while LLM2Vec~\cite{behnamghader2024llmvec}
shows that bidirectional attention can become beneficial after proper adaptation. More discussion is provided in Section~\ref{app:bidirectional_mask_discussion}.
Motivated by these findings, we revisit the causal ordering imposed on memory
slots in soft context compression. Under causal masking, memory slots are forced into an unnecessary
left-to-right order. This constraint is natural for autoregressive generation, but not for soft
context compression. Since memory slots are latent representations 
rather than tokens to be generated, they should communicate with each other
and cooperate to allocate different aspects of the query-relevant context.

Figure~\ref{fig:main} (Right) illustrates our attention-mask design. In detail, we introduce a bidirectional flow
among memory slots while preserving the causal structure of the original
query-context sequence. Let $L=|Q|+|C|$ denote the length of the
query-context prefix and $K$ denote the number of memory slots. The
encoder input is arranged as $[Q, C, M]$, and its attention mask is defined
in Eq.~\ref{eq:bidirectional-mask}:
\begin{equation}
\label{eq:bidirectional-mask}
\mathbf{A}
=
\begin{bmatrix}
\mathrm{LowerTri}(L) & \mathbf{0}_{L\times K} \\
\mathbf{1}_{K\times L} & \mathbf{1}_{K\times K}
\end{bmatrix},
\end{equation}
where $\mathrm{LowerTri}(L)$ is an $L \times L$ binary lower-triangular matrix
that encodes causal attention over the query-context prefix. The lower-right block makes memory slots
bidirectional, allowing them to exchange information and form a coordinated
set of compression slots. We implement the corresponding bidirectional mask using
FlexAttention for efficient custom attention masking.\footnote{\url{https://pytorch.org/blog/flexattention/}}

\begin{table*}[t]
    \caption{Overall performance on four question-answering benchmarks and
    RULER-NIAH. All rows use Mistral-7B except the final two \method{} rows,
    which use Qwen3-4B. SeleCom-82$\times$$^{*}$ denotes the best-performing
    configuration reported in the original paper, and $^\triangle$ marks
    query-aware methods. Bold and underlined numbers denote the best and
    second-best results within each Mistral-7B compression setting.}
    \label{tab:main_results}
    \centering
    \small
    \setlength{\tabcolsep}{2pt}
    \begin{tabular}{llccccccccc}
        \toprule
        Type & Model & \multicolumn{2}{c}{2Wiki} & \multicolumn{2}{c}{HotpotQA} & \multicolumn{2}{c}{MuSiQue} & \multicolumn{2}{c}{NQ} & NIAH \\
        \cmidrule(lr){3-4} \cmidrule(lr){5-6} \cmidrule(lr){7-8} \cmidrule(lr){9-10} \cmidrule(lr){11-11}
        & & ACC & LLM Judge & ACC & LLM Judge & ACC & LLM Judge & ACC & LLM Judge & ACC \\
        \midrule
        \multicolumn{11}{c}{RAG (No Compression)} \\
        \midrule
        & Mistral-7B & 0.7654 & 0.7088 & 0.7751 & 0.8249 & 0.5333 & 0.5085 & 0.7905 & 0.7671 & 0.886 \\
        & Qwen3-4B & 0.7927 & 	0.6236 &	0.8164& 	0.8220 &	0.3814 &	0.2880 &	0.8007 	& 0.7405	&1.0 \\
        \midrule
        \multicolumn{11}{c}{Mistral-7B: 4$\times$ Compression} \\
        \midrule
        \multirow{3}{*}{\shortstack{Hard\\compression}}
        & LLMLingua2 & 0.3905 & 0.3379 & 0.3612 & 0.4848 & 0.1614 & 0.2284 & 0.5081 & 0.6121 & 0.790 \\
        & LongLLMLingua & 0.4436 & 0.2741 & 0.4254 & 0.4898 & 0.1465 & 0.1825 & 0.6084 & 0.6630 & 0.128 \\
        & Provence$^\triangle$ & \underline{0.6131} & \underline{0.6279} & \underline{0.6624} & 0.6415 & \textbf{0.3620} & \textbf{0.3471} & \underline{0.7522} & 0.7318 & 0.522 \\
        \midrule
        \multirow{4}{*}{\shortstack{Soft\\compression}}
        & ICAE & 0.3941 & 0.4282 & 0.4043 & 0.5352 & 0.1820 & 0.2561 & 0.4686 & 0.5722 & 0.782 \\
        & PISCO & 0.5884 & 0.5740 & 0.6551 & \underline{0.6691} & 0.3194 & 0.3029 & 0.7368 & \underline{0.7921} & \underline{0.960} \\
        & RAM$^\triangle$ & 0.4614 & 0.5079 & 0.4510 & 0.5859 & 0.1994 & 0.2896 & 0.5075 & 0.6269 & 0.948 \\
        & \method{} & \textbf{0.6280} & \textbf{0.6662} & \textbf{0.6937} & \textbf{0.7045} & \underline{0.3359} & \underline{0.3277} & \textbf{0.7902} & \textbf{0.8185} & \textbf{0.972} \\
        \midrule
        \multicolumn{11}{c}{Mistral-7B: 32$\times$ Compression} \\
        \midrule
        \multirow{3}{*}{\shortstack{Hard\\compression}}
        & LLMLingua2 & 0.3318 & 0.2465 & 0.2241 & 0.2718 & 0.0368 & 0.0621 & 0.1945 & 0.2665 & 0.382 \\
        & LongLLMLingua & 0.4120 & 0.2139 & 0.3403 & 0.3727 & 0.0869 & 0.1051 & 0.5361 & 0.5724 & 0.000 \\
        & Provence$^\triangle$ & 0.5017 & 0.4908 & 0.4358 & 0.4752 & 0.1444 & 0.1601 & 0.5736 & 0.6362 & 0.486 \\
        \midrule
        \multirow{6}{*}{\shortstack{Soft\\compression}}
        & ICAE & 0.3403 & 0.3721 & 0.3161 & 0.4410 & 0.0761 & 0.1353 & 0.3856 & 0.5079 & 0.308 \\
        & PISCO & \underline{0.5369} & \underline{0.5306} & \underline{0.5683} & \underline{0.6140} & \textbf{0.2731} & \textbf{0.2842} & 0.6731 & 0.7459 & \underline{0.578} \\
        & RAM$^\triangle$ & 0.2723 & 0.2960 & 0.1943 & 0.2637 & 0.0294 & 0.0720 & 0.1615 & 0.2447 & 0.000 \\
        & SeleCom$^\triangle$ & 0.2894 & 0.3331 & 0.3445 & 0.4706 & 0.0856 & 0.1593 & 0.6531 & 0.7604 & 0.000 \\
        & SeleCom-82$\times$$^{*\triangle}$ & 0.3410 & 0.3883 & 0.4191 & 0.5622 & 0.1477 & 0.2284 & \underline{0.7315} & \textbf{0.8229} & 0.000 \\
        & \method{} & \textbf{0.5650} & \textbf{0.5479} & \textbf{0.6197} & \textbf{0.6586} & \underline{0.2656} & \underline{0.2578} & \textbf{0.7389} & \underline{0.7862} & \textbf{0.612} \\
        \midrule
        \multicolumn{11}{c}{Qwen3-4B: \method{} Only} \\
        \midrule
        \multirow{2}{*}{\shortstack{Soft\\compression}}
        & \method{}-4$\times$ & 0.6102 & 0.6441 & 0.6900 & 0.7489 & 0.3893 & 0.3827 & 0.7249 & 0.8226 & 0.948 \\
        & \method{}-32$\times$ & 0.5591 & 0.5187 & 0.6349 & 0.6582 & 0.2971 & 0.2958 & 0.7013 & 0.7880 & 0.634 \\
        \bottomrule
    \end{tabular}
\end{table*}

\section{Experiments}

\subsection{Experimental Setup}

\subsubsection{Training Data and Details.}
We train \method{} in two stages. In Stage I, we construct a query-rich
training set from the CLaRa multi-stage pretraining
corpus.\footnote{\url{https://huggingface.co/datasets/apple/CLaRa_multi_stage/tree/main/pretraining}}
To support same-context hard negatives, we keep only contexts paired with
at least eight distinct queries, yielding approximately 160K training
instances. In Stage II, following~\citet{rau2025context}\footnote{\url{https://huggingface.co/datasets/dmrau/multi_qa}}, we perform instruction tuning on approximately 453K instances.

In Stage I, both the compressor $f_{\phi}$ and decoder $p_{\theta}$ are trained with LoRA. We use Mistral-7B\footnote{\url{https://huggingface.co/mistralai/Mistral-7B-Instruct-v0.2}} and Qwen3-4B\footnote{\url{https://huggingface.co/Qwen/Qwen3-4B-Instruct-2507}} as the backbone LLMs.
Following~\citet{ge2024incontext}, each document
is split into 256-token chunks. For a context chunk of length $L$, the
compressor produces memory states at compression ratios
$r \in \{4,8,16,32, 64, 128\}$, with the number of memory slots determined by the
target ratio. We apply the memory-context contrastive loss
$\mathcal{L}_{\mathrm{cl}}$ with weight $\lambda = 0.1$. In Stage II, we initialize
from the Stage I checkpoint, freeze the compressor $f_{\phi}$, and continue
training only the decoder $p_{\theta}$. Additional details on data and
training configurations are provided in Appendix~\ref{app:traing_details}.

\subsubsection{Evaluation Benchmarks.}
We evaluate on four long-document question answering benchmarks:
Natural Questions~\cite{kwiatkowski-etal-2019-natural},
HotpotQA~\cite{yang-etal-2018-hotpotqa},
MuSiQue~\cite{trivedi-etal-2022-musique}, and
2WikiMultiHopQA~\cite{ho-etal-2020-constructing}. These datasets cover
both single-hop and multi-hop reasoning. We also include
RULER-NIAH~\cite{hsieh2024ruler}, a synthetic needle-in-a-haystack
benchmark in which a 7-digit numerical needle is inserted into five
documents. This benchmark directly tests whether the memory slots
$M=f_{\phi}(C,Q)$ retain query-relevant evidence from the original context
$C$ under long-context retrieval.

\subsubsection{Metrics.}
We report Cover Exact Match, denoted as accuracy (ACC), and LLM-as-a-judge scores. Unlike standard Exact Match and F1 metrics, Cover-EM measures whether the generated answer contains the gold answer span. In addition, we use DeepSeek-V4-Flash as an LLM-as-a-judge evaluator to better assess the semantic correctness of model predictions. The full prompt template is provided in Appendix~\ref{app:llm_judge_prompt}, and further analysis on the motivation for using LLM-as-a-judge instead of EM/F1 is presented in Appendix~\ref{app:why_llm_judge}.

\subsubsection{Baselines.}
We compare \method{} with both hard and soft context compression methods.
For hard compression, we evaluate LLMLingua-2
\cite{pan-etal-2024-llmlingua}, LongLLMLingua
\cite{jiang-etal-2024-longllmlingua}, and the query-aware Provence
\cite{chirkova2025provence}. For soft compression, we include
ICAE~\cite{ge2024incontext}, PISCO~\cite{louis-etal-2025-pisco}, and the
query-aware RAM~\cite{Tang2026ReadAH} and SeleCom~\cite{Liu2026SeleCom}.
For all evaluated methods, we use Mistral-7B as the  decoder. Additionally, we also report model performance under the settings of
the original prompt, corresponding to standard RAG without context compression.

\subsection{Main Results}

\subsubsection{Overall Performance.}
Table~\ref{tab:main_results} summarizes results on four long-document QA benchmarks and RULER-NIAH. Overall, \method{} consistently outperforms prior compression baselines across compression ratios. At 4$\times$ compression, \method{} achieves the best overall QA performance, substantially improving multi-hop reasoning performance. Notably, under  llm judge score, \method{} even surpasses the uncompressed full-context baseline on several benchmarks. Even under much tighter memory budgets (32$\times$ compression), \method{} remains highly competitive, consistently outperforming strong baselines while maintaining robust retrieval performance on RULER-NIAH.

Hard compression methods generally preserve sparse lexical evidence but often hurt QA performance, especially under high compression ratios. In contrast, soft compression methods retain richer semantic information and are more stable on QA benchmarks, but their performance drops noticeably as memory budgets become tighter. In comparison, \method{} maintains strong performance across both retrieval and reasoning tasks, suggesting that query-aware memory construction makes more effective use of limited memory capacity.

\begin{table}[t]
\caption{Accuracy (ACC) of \method{} across different compression ratios.}
    \label{tab:backbone_results}
    \centering
    \small
    \setlength{\tabcolsep}{2pt}

    \begin{tabular}{lcccc}
        \toprule
        Model & 2Wiki & HotpotQA & MuSiQue & NQ  \\
        \midrule
        Mistral-7B-8  $\times$ & 0.5863 & 0.6437 & 0.2826 & 0.7524 \\
        Mistral-7B-16  $\times$ & 0.5988 & 0.6411 & 0.2627 & 0.7524 \\
        Mistral-7B-64 $\times$  & 0.5724 & 0.5925  & 0.2466  & 0.7223     \\
         Mistral-7B-128 $\times$  & 0.5756 & 0.5975  & 0.2594  & 0.7294     \\
        \midrule
        Qwen3-4B-8$\times$     & 0.5833 & 0.6578 & 0.3434 & 0.7010 \\
        Qwen3-4B-16$\times$    & 0.5540 &0.6366 &0.2974 &0.6997  \\
        Qwen3-4B-64$\times$   & 0.5069 &0.5864  &0.2378  &0.6777   \\
        Qwen3-4B-128$\times$    & 0.4941&0.5682 &0.2304 &0.6640\\
        \bottomrule
    \end{tabular}
\end{table}

\subsubsection{Additional Compression Ratios.}
We further evaluate \method{} under increasingly aggressive compression ratios to assess its robustness under highly constrained memory budgets. As shown in Table~\ref{tab:backbone_results}, \method{} maintains relatively stable performance as the compression ratio increases from $8\times$ to $128\times$ across both Mistral-7B and Qwen3-4B. In particular, for Qwen3-4B, increasing the compression ratio to $128\times$ results in an average performance drop of only 0.0822 compared with $8\times$ compression. These results demonstrate that \method{} exhibits graceful degradation even under extremely high compression ratios, retaining a substantial portion of its performance while reducing the memory footprint by up to $128\times$.

\subsection{Ablation Study}

\begin{table}[t]
    \caption{Ablation study on Stage I and Stage II training components.}
    \label{tab:training_ablation}
    \centering
    \small
    \setlength{\tabcolsep}{5pt}
    \resizebox{0.99\linewidth}{!}{
    \begin{tabular}{lccccc}
        \toprule
        Contrastive & Bi-Mask & 2Wiki & HotpotQA & MuSiQue & NQ \\
        \midrule
        \multicolumn{6}{l}{\textit{Stage I}} \\
        $\times$ & $\checkmark$ & 0.2785 & 0.2440 & 0.0709 & 0.2560 \\
        $\times$ & $\times$ & 0.2852 & 0.2447 & 0.0861 & 0.2665 \\
        $\checkmark$ & $\times$ & 0.2627 & 0.1492 & 0.0203 & 0.1160 \\
        $\checkmark$ & $\checkmark$ & \textbf{0.3761} & \textbf{0.5211} & \textbf{0.2280} & \textbf{0.6197} \\
        \midrule
        \multicolumn{6}{l}{\textit{Stage II}} \\
        \multicolumn{2}{l}{Ours} & 0.5591 & \textbf{0.6349} & \textbf{0.2971} & \textbf{0.7013} \\
        \multicolumn{2}{l}{w/ encoder} & \textbf{0.5782} & 0.6211 & 0.2950 & 0.6697 \\
        \multicolumn{2}{l}{w/o Stage I checkpoint} & 0.3932 & 0.2187 & 0.0360 & 0.2319 \\
        \bottomrule
    \end{tabular}
    }
\end{table}

\begin{figure}[!tb]
    \centering
    \includegraphics[width=0.98\linewidth]{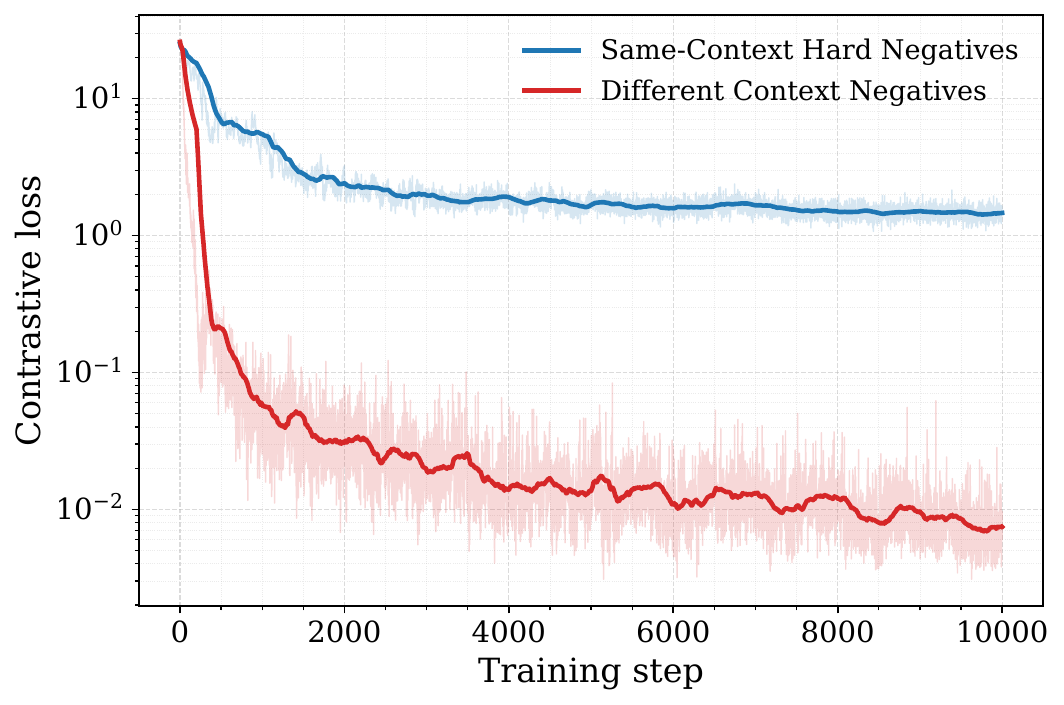}
    \caption{Optimization dynamics for same-context and different-context negatives.}
    \Description{Optimization dynamics for same-context and different-context negatives.}
    \label{fig:loss_compare}
\end{figure}

\subsubsection{Interaction between Contrastive Learning and Bidirectional Memory.}
Unless otherwise specified, all ablation experiments are conducted under a
32$\times$ compression ratio and report ACC. Table~\ref{tab:training_ablation} shows the effects of the key training components of 
\method{}. \textbf{In Stage I}, contrastive learning and the bidirectional memory mask
are strongly complementary. Without contrastive learning, the choice of memory
mask has only a limited effect. Once contrastive learning is applied, however,
using a causal mask leads to a substantial performance drop, whereas combining
contrastive learning with bidirectional memory achieves the best results
across all datasets. This suggests that contrastive supervision requires
non-causal interactions among memory slots to effectively allocate limited
capacity and preserve query-relevant information. \textbf{Stage II} results further
show that the Stage I checkpoint is critical for downstream answer generation,
while updating the encoder during Stage II provides no consistent benefit over
freezing the compressor.

\begin{table}[t]
    \caption{Ablation study on same-context hard negatives and different-context negatives.}
    \label{tab:negative_type_ablation}
    \centering
    \small
    \setlength{\tabcolsep}{5pt}
    \resizebox{0.99\linewidth}{!}{
    \begin{tabular}{lcccc}
        \toprule
        Negative Type & 2Wiki & HotpotQA & MuSiQue & NQ \\
        \midrule
        Different-Context 
        & 0.3681 & 0.4947 & 0.2201 & 0.6047 \\
        Same-Context 
        & \textbf{0.3761} & \textbf{0.5211} & \textbf{0.2280} & \textbf{0.6197} \\
        \bottomrule
    \end{tabular}
    }
\end{table}

\subsubsection{Effect of Same-Context Hard Negatives.}
We next test whether adding negatives from unrelated contexts improves the
contrastive objective. In conventional contrastive learning, more negatives
often provide stronger discrimination. In our setting, however, the key
challenge is not to separate different documents, but to distinguish which
evidence should be preserved for different queries over the same document.
Different-context negatives can therefore be separated using coarse
document-level cues and provide a weaker signal for query-specific
compression. As shown in Figure~\ref{fig:loss_compare}, the optimization
dynamics also indicate that such negatives contribute less useful training
signal than same-context hard negatives. Consistent with this observation,
Table~\ref{tab:negative_type_ablation} shows that same-context negatives
outperform different-context negatives on all four datasets.

\subsection{Deep Analysis of Selective Compression}

\begin{table}[t]
    \caption{ACC with the true downstream query or a general-purpose query at
    32$\times$ compression ratios.}
    \label{tab:general_query_ablation}
    \centering
    \setlength{\tabcolsep}{4pt}
    \resizebox{0.99\linewidth}{!}{
    \begin{tabular}{lcccc}
        \toprule
        Method & 2Wiki & HotpotQA & MuSiQue & NQ \\
        \midrule
        \method{} w/ true query
        & \textbf{0.5591} & \textbf{0.6349} & \textbf{0.2971} & \textbf{0.7013} \\
        \method{} w/ general query
        & 0.3786 & 0.4607 & 0.2098 & 0.5597 \\
        \bottomrule
    \end{tabular}
    }
\end{table}

\subsubsection{General Compression without Downstream Queries.}
Although \method{} conditions compression on a query, its context-preservation
objective also encourages the memory to retain information from the source
context. At evaluation time, we replace the true query with the generic
instruction \emph{``Please preserve all potentially useful information from the
following context.''} As shown in Table~\ref{tab:general_query_ablation},
\method{} with this general-purpose query still achieves strong performance
under 32$\times$ compression, consistently outperforming the query-agnostic
baselines ICAE and LLMLingua2 across all four datasets. While access to the true query yields the strongest results, the general-query
variant retains substantial performance, enabling the resulting memories to be
precomputed offline, stored, and reused for different downstream queries.

\subsubsection{Efficiency Analysis.}

\begin{table}[t]
\caption{Efficiency comparison of \method{} across different compression ratios.}
\label{tab:online_offline_efficiency}
\centering
\small
\setlength{\tabcolsep}{6pt}

\begin{tabular}{lrrrr}
\toprule
\multicolumn{5}{c}{GFlops} \\
\midrule
Compression ratio & 4$\times$   & 16$\times$   & 32$\times$   & 64$\times$   \\
\midrule
Full Context & \multicolumn{4}{c}{10031} \\
Ours & 14074 & 10800 & 10256 & 9985 \\
\midrule
\multicolumn{5}{c}{Time/sample (s)} \\
\midrule
Compression ratio & 4$\times$   & 16$\times$   & 32$\times$   & 64$\times$   \\
\midrule
Full Context & \multicolumn{4}{c}{0.053} \\
Ours & 0.026 & 0.013 & 0.012 & 0.011 \\
 \midrule
\multicolumn{5}{c}{Speedup} \\
        \midrule
        \method{}
        & 2.04$\times$ & 4.08$\times$ & 4.42$\times$ & \textbf{4.82$\times$} \\
\bottomrule
\end{tabular}
\end{table}

Building on this offline compression setting, we next evaluate the inference efficiency of \method{} on a single H100 GPU using a setup with a 128-token query, five 256-token documents, and forced generation of 32 tokens.   We report only the decoder-side answer generation time, excluding retrieval and context compression overhead, to isolate the efficiency of downstream generation. Table~\ref{tab:online_offline_efficiency} reports both theoretical GFlops and measured latency in seconds per sample under different compression ratios. Compared with the full-context baseline, \method{} substantially reduces inference latency as the compression ratio increases, achieving up to a 4.8$\times$ speedup at 64$\times$ compression while maintaining comparable computational cost in terms of GFlops.

\begin{figure}[!tb]
    \centering
    \includegraphics[width=0.99\linewidth]{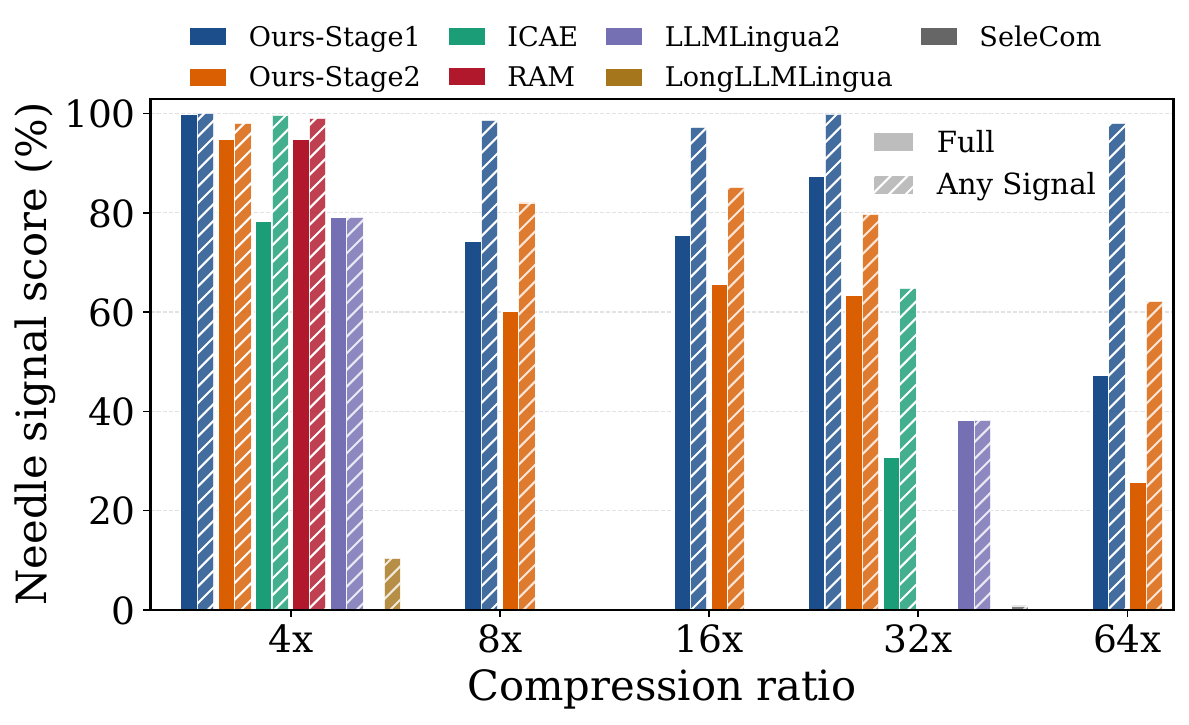}
    \caption{Needle-signal retention under different compression ratios.}
    \Description{Needle-signal retention under different compression ratios.}
    \label{fig:needle_signal_lineplot}
\end{figure}

\subsubsection{Needle-Signal Retention.} Using Qwen3-4B as the backbone LLM, we evaluate RULER-NIAH to assess whether the compressor preserves retrieval-critical information. We report two metrics: \textit{Full}, which requires the complete 7-digit needle, and \textit{Any Signal}, which counts either the full needle or a recognizable digit substring. Thus, \textit{Any Signal} captures partial retention of query-relevant evidence even when exact recovery fails.

As shown in Figure~\ref{fig:needle_signal_lineplot}, existing hard compression and soft compression methods exhibit near-zero performance  under aggressive compression, due to their limited query-aware evidence selection. Query-agnostic soft compression retains more evidence at lower compression ratios but also degrades substantially as the memory budget shrinks.

In contrast, \method{} preserves substantially more query-relevant evidence under extreme compression. Stage I achieves a 0.98 \textit{Any Signal} score at 64$\times$ compression, demonstrating reliable needle localization despite the severe memory reduction. The gap between \textit{Full} and \textit{Any Signal} indicates that remaining errors are often due to partial corruption rather than complete loss of the relevant evidence. These results highlight the effectiveness of query-aware contrastive compression in retaining retrieval-critical information under extreme compression.

\section{Related Work}
\paragraph{Hard Prompt Compression.}
One line of work studies compressing the original context by removing
unimportant text tokens using a lightweight auxiliary model, while
preserving readable prompts for direct use by off-the-shelf LLMs. Representative approaches include
\textsc{LLMLingua}~\cite{jiang-etal-2023-llmlingua},
\textsc{LongLLMLingua}~\cite{jiang-etal-2024-longllmlingua},
and \textsc{LLMLingua-2}~\cite{pan-etal-2024-llmlingua}. Provence
\cite{chirkova2025provence} formulates context pruning as sequence labeling
and dynamically retains relevant passages. These methods filter the original
context, making them prone to removing query-relevant or otherwise
critical information during aggressive compression.
DAC~\cite{zhao-etal-2025-dac} further incorporates attention scores and
dynamic entropy changes to preserve salient tokens. 
This reliance on full attention incurs additional overhead, since modern
FlashAttention kernels~\cite{dao2024flashattention,shah2024flashattention} do not save full attention matrices.

\paragraph{Soft Context Compression.}
Unlike hard compression methods, soft compression encodes long contexts into
continuous latent representations and feeds them to a decoder for generation. \citet{ge2024incontext} compress
context into learned memory slots through autoencoding and language modeling
objectives, while \citet{li2025500xcompressor} explore extreme compression by
using KV caches as compressed representations. \citet{dai2025pretraining}
pretrain a decoupled compressor with embedding-based memory using reconstruction
and completion tasks. More recent works reduce the reliance on explicit reconstruction pretraining~\cite{louis-etal-2025-pisco,liu2026iclr-autoencodingfree}.
 \citet{lin2025refrag} mix compressed chunk embeddings with standard token inputs
for efficient RAG decoding, \citet{he2025clara} jointly optimize retrieval and
generation over compressed  memories. Closest
to our setting, \citet{louis2026oscar} and \citet{Liu2026SeleCom} condition
compression on the input query to reduce irrelevant information.  In summary, existing soft compressors largely rely on decoder-side objectives,
whereas \method{} explicitly regularizes compressed memories to be both
context-preserving and query-guided.

\paragraph{Bidirectional Masking}
\label{app:bidirectional_mask_discussion}
Prior work suggests that bidirectional masks can improve representations by
relaxing the information bottleneck imposed by causal attention, but the gains
are not automatic~\cite{li2023label,li2024bellm,behnamghader2024llmvec,muennighoff2025generative,lee2025nvembed}. 
These studies show that access to right context is useful for token- and
sequence-level embeddings when paired with suitable adaptation, while naive
mask changes can be less reliable. Similar findings appear in context
compression, where full-context attention helps compressed vectors or anchor
tokens aggregate global information~\cite{feldman2025simplecontextcompression,liu2026iclr-autoencodingfree}.
This motivates our use of bidirectional communication among memory slots, which
are latent representations rather than autoregressively generated tokens.
\section{Conclusion}
In this paper, we introduce \method{}, a query-guided selective context compression framework. Through contrastive compressor learning, same-context hard negatives, and bidirectional memory interaction, \method{} effectively preserves query-relevant evidence under compact memory budgets. Extensive experiments on long-document QA benchmarks and RULER-NIAH demonstrate that \method{} consistently outperforms prior  compression methods across compression ratios.

\appendix

\definecolor{blue-color}{RGB}{40,166,189}
\definecolor{pink-color}{RGB}{237,46,104}
\definecolor{dark-grey-color}{RGB}{79,91,102}
\providecommand{\param}[1]{
\textcolor{black}{\textnormal{\detokenize{#1}}}
}
\providecommand{\prompttext}[1]{\textcolor{black}{#1}}

\newtcolorbox[list inside=prompt,auto counter,number within=section]{prompt}[1][]{
    colbacktitle=black!80,
    colframe=black!80,
    coltitle=white,
    fontupper=\footnotesize,
    boxsep=5pt,
    left=0pt,
    right=0pt,
    top=0pt,
    bottom=0pt,
    boxrule=1pt,
    enhanced,
    breakable,
    skin first=enhanced,
    skin middle=enhanced,
    skin last=enhanced,
    #1,
}

\section{Implementation details}
\label{app:traing_details}
The training hyperparameters are summarized in
Table~\ref{tab:training_hyperparameters}. Stage I training can be completed
within 10 hours on a single H100 GPU. We set the micro-batch size to 8, which
matches one same-context group containing eight distinct queries. Table~\ref{tab:same_context_contrastive_examples} provides an example of such a same-context contrastive group. Stage II
training finishes within 5 hours on four H100 GPUs. The training time remains
similar across different compression ratios. Overall, \method{} is efficient
to train.

\section{Prompt Templates}
\label{app:prompt_templates}
\subsection{LLM-as-a-Judge Evaluation Prompt}
\label{app:llm_judge_prompt}

For LLM-as-a-judge evaluation, we use DeepSeek-V4-Flash as the evaluator.
Each example is formatted as a single user message, and the judge is asked to
return only a binary decision. The prompt template is shown in
Prompt~\ref{prompt:llm_judge}.

\begin{prompt}[title={Prompt \thetcbcounter: LLM-as-a-Judge Evaluation Prompt}, label=prompt:llm_judge]
\prompttext{
You are an answer correctness judge. Given a question, a reference answer, and a model's prediction, determine if the prediction is correct.\\
A prediction is correct if it conveys the same meaning as the reference answer, even if the wording is different. Minor differences in formatting, articles, or extra context are acceptable as long as the core answer is correct.\\
\\
Question: \param{{question}}\\
Reference Answer: \param{{answer}}\\
Model Prediction: \param{{prediction}}\\
\\
Is the model's prediction correct? Reply with only ``Yes'' or ``No''.
}
\end{prompt}

\subsection{Training Data Prompt Templates}
\label{app:training_prompt_templates}

During training, \method{} uses separate templates for the compression-side
input and the decoder-side answer generation input. The compression input
places the query between two \texttt{\textless QUERY\textgreater} markers,
followed by the source document, and appends memory tokens after tokenization.
The decoder-side prompt receives the compressed memory slots as the background
and generates the target answer.

\begin{prompt}[title={Prompt \thetcbcounter: Query-Conditioned Compression Input}, label=prompt:encoder_training_template]
\prompttext{
\texttt{\textless ENC\textgreater}\texttt{\textless BOS\textgreater}\texttt{\textless QUERY\textgreater}\\
Query:\\
\param{{query}}\texttt{\textless QUERY\textgreater}Document:\\
\param{{document}}\texttt{\textless EOS\textgreater}\\
\texttt{\textless MEM\textgreater}\ldots\texttt{\textless MEM\textgreater}
}
\end{prompt}

\begin{prompt}[title={Prompt \thetcbcounter: Decoder QA Template}, label=prompt:decoder_qa_template]
\prompttext{
\textbf{System:}\\
You are a helpful assistant. Your task is to extract relevant information from provided documents and to answer questions.\\
\\
\textbf{User:}\\
Background:\\
\param{{compressed memory slots}}\\
\\
Question:\param{{query}}\\
\\
\textbf{Assistant:}\\
\param{{answer}}
}
\end{prompt}

\begin{table}[t!]
    \caption{Training hyperparameters.}
    \label{tab:training_hyperparameters}
    \small
    \centering
    \begin{tabular}{@{}ll@{}}
    \toprule
    \textbf{Hyperparameter} & \textbf{Value} \\
    \midrule
    Batch size & 128 \\
    Micro batch size per GPU & 8 \\
    Gradient accumulation steps & 16 \\
    Learning rate & $1\times10^{-4}$ \\
    LR scheduler & cosine with minimum LR  \\
    Optimizer & DeepSpeed FusedAdam \\
    Epochs & 1 \\
    Max tokens for generation & 128 \\
    Document max tokens & 256 \\
    LoRA layers & All-linear \\
    LoRA rank & 16 \\
    LoRA dropout & 0.1 \\
    LoRA alpha & 32 \\
    Warmup ratio & 0.03 \\
    Number of questions per context & 8 \\
    \bottomrule
    \end{tabular}
\end{table}

\section{Why Use LLM-as-a-Judge?}
\label{app:why_llm_judge}

\paragraph{Biases of Automatic Metrics.}
Prior work commonly reports exact match EM, F1, and ACC as the main evaluation metrics~\cite{li2025500xcompressor,he2025clara,liu2026iclr-autoencodingfree}. 
Given a prediction $\hat{y}$ and a reference answer $y$, let $n(\cdot)$ denote the evaluation normalization, which lowercases text, removes punctuation and articles, maps boolean labels to yes/no, and fixes whitespace. Let $T(\cdot)$ denote the token multiset after normalization and $S(\cdot)$ denote the corresponding token set. These metrics can be written as:
\begin{align}
\mathrm{EM}(\hat{y}, y)
&=
\mathbb{I}\left[n(\hat{y}) = n(y)\right],
\\
\mathrm{F1}(\hat{y}, y)
&=
\frac{2\,\mathrm{Prec}(\hat{y}, y)\,\mathrm{Rec}(\hat{y}, y)}
{\mathrm{Prec}(\hat{y}, y)+\mathrm{Rec}(\hat{y}, y)},
\\
\mathrm{ACC}(\hat{y}, y)
&=
\mathbb{I}\left[S(y)\subseteq S(\hat{y})\right],
\end{align}
where $\mathrm{Prec}$ and $\mathrm{Rec}$ are computed by multiset token overlap between $T(\hat{y})$ and $T(y)$.
These definitions show that the metrics are still largely based on lexical overlap with the reference answer.
EM is especially strict, as it requires the model output to exactly match the gold label.
In our experiments, we observed that EM is strongly affected by the output length distribution in the training data.
When the training answers are short, the model also tends to produce shorter responses, which can artificially improve EM and F1.
However, this behavior often reduces answer completeness, weakens the model's reasoning ability, and may even lower ACC when the model fails to provide enough supporting information.

\begin{table*}[t!]
    \caption{A same-context example from the contrastive data. One context about the history of Azerbaijani press supports eight different queries and answers.}
    \label{tab:same_context_contrastive_examples}
    \scriptsize
    \centering
    \begin{tabular}{@{}p{0.08\textwidth}p{0.39\textwidth}p{0.45\textwidth}@{}}
    \toprule
    \multicolumn{3}{@{}p{0.95\textwidth}@{}}{\textit{Shared context.}
    This document describes the history of Azerbaijani press. Hasan bey Zardabi founded publishing in Azerbaijan and sought permission to publish the native-language newspaper ``Akinchi'' under the Tsar's colonial system. The newspaper drew a wide response in the Caucasus, and Zardabi served as its publisher, editor, corrector, and typesetter. From 1875 to 1877, 56 issues were published. Zardabi distributed the newspaper gratuitously to arouse public interest and used it to promote national enlightenment, improve peasants' agricultural literacy, and support national unity. During the Russo-Turkish War of 1877--1878, Tsarist censorship prohibited political discussion, and in 1877 a decree ordered the newspaper's liquidation.} \\
    \midrule
    \textbf{ID} & \textbf{Query} & \textbf{Answer} \\
    \midrule
    1 & Who is considered the founder of publishing in Azerbaijan? & Hasan bey Zardabi is considered the founder of publishing in Azerbaijan. \\
    2 & During which period was the newspaper ``Akinchi'' published? & The newspaper ``Akinchi'' was published during 1875--1877. \\
    3 & How many issues of the ``Akinchi'' newspaper were published between 1875 and 1877? & 56 issues. \\
    4 & What role did Hasan bey Zardabi play in the publication of the ``Akinchi'' newspaper? & Hasan bey Zardabi was the founder, publisher, editor, corrector, and typesetter of the ``Akinchi'' newspaper. \\
    5 & What was the main goal of Hasan bey Zardabi in publishing the ``Akinchi'' newspaper? & His main goal was the propagation of ideas of the nation's enlightenment. \\
    6 & What major event caused the situation of the ``Akinchi'' newspaper to sharply complicate in 1877? & The beginning of the Russo-Turkish War of 1877--1878. \\
    7 & What restriction was imposed by the Tsar's censorship on the ``Akinchi'' newspaper during the Russo-Turkish War? & Tsar's censorship prohibited the newspaper from touching upon political issues. \\
    8 & What decree was issued in 1877 regarding the ``Akinchi'' newspaper? & A decree ordering the liquidation of the newspaper was issued. \\
    \bottomrule
    \end{tabular}
\end{table*}

\begin{table*}[t!]
    \caption{Qualitative examples where lexical ACC marks the Qwen3-4B original output as correct because the gold answer is mentioned, while LLM-as-a-Judge rejects it due to refusal, contradiction, or over-qualification. In contrast, \method{} produces a direct answer that is accepted by both ACC and LLM-as-a-Judge.}
    \label{tab:llm_judge_cases}
    \small
    \centering
    \begin{tabular}{@{}c p{0.78\textwidth}@{}}
    \toprule
    \multicolumn{1}{r}{\textit{Question}} & What contest did Drew Barry's younger brother win in 1996? \\
    \multicolumn{1}{r}{\textit{Gold Answer}} & the 1996 NBA Slam Dunk Contest \\
    \multicolumn{1}{r}{Qwen3-4B} & \cellcolor{gray!15} There is no information in the provided documents indicating that Drew Barry's younger brother won any contest in 1996. In fact, the documents mention that Brent Barry won the \uline{1996 NBA Slam Dunk Contest}, but Brent is not Drew Barry's younger brother. However, the documents do not state that Brent won a contest in 1996 as a younger brother of Drew. \color{red}{(mentions the answer but rejects it)} \\
    \multicolumn{1}{r}{\method{}} & Drew Barry's younger brother, Brent, won the NBA Slam Dunk Contest in 1996. \\
    \cmidrule{1-2}
    \multicolumn{1}{r}{\textit{Question}} & Who is the actor who plays the title character in The Bourne Betrayal? \\
    \multicolumn{1}{r}{\textit{Gold Answer}} & Matt Damon \\
    \multicolumn{1}{r}{Qwen3-4B} & \cellcolor{gray!15} The actor who plays the title character in The Bourne Betrayal is \uline{Matt Damon}. However, it is important to clarify: The Bourne Betrayal is a novel by Eric Van Lustbader, not a film. Matt Damon does not appear in the novel, as it is a book, not a movie. Therefore, there is no actor who plays the title character in the novel The Bourne Betrayal. \color{red}{(self-contradictory final answer)} \\
    \multicolumn{1}{r}{\method{}} & Matt Damon plays the title character in The Bourne Betrayal. \\
    \cmidrule{1-2}
    \multicolumn{1}{r}{\textit{Question}} & Who is the mother-in-law of Fredegund? \\
    \multicolumn{1}{r}{\textit{Gold Answer}} & Aregund \\
    \multicolumn{1}{r}{Qwen3-4B} & \cellcolor{gray!15} The provided documents do not mention the mother-in-law of Fredegund. Fredegund was the queen consort of Chilperic I, who was the son of Clotaire I and Queen \uline{Aregund}. Therefore, Aregund would be the mother-in-law of Fredegund. However, this information is not explicitly stated in the given documents. Strictly from the provided text, there is no information about Fredegund's mother-in-law. \color{red}{(mentions the answer but refuses)} \\
    \multicolumn{1}{r}{\method{}} & Chilperic I was the son of Fredegund's mother-in-law, Queen Aregund. \\
    \cmidrule{1-2}
    \multicolumn{1}{r}{\textit{Question}} & Who sings Lean on Me in the movie? \\
    \multicolumn{1}{r}{\textit{Gold Answer}} & Big Daddy Kane \\
    \multicolumn{1}{r}{Qwen3-4B} & \cellcolor{gray!15} The document does not specify who sings ``Lean on Me'' in the movie Lean on Me. However, it does mention that the song was adapted by \uline{Big Daddy Kane} in a hip-hop form for the film. While this indicates a version of the song was used in the film, it does not confirm whether Big Daddy Kane's version was the one performed in the movie. \color{red}{(over-qualified and non-committal)} \\
    \multicolumn{1}{r}{\method{}} & Big Daddy Kane sings ``Lean on Me'' in its hip-hop form in the movie Lean on Me. \\
    \cmidrule{1-2}
    \multicolumn{1}{r}{\textit{Question}} & When did Season 3 of The Expanse start on Netflix? \\
    \multicolumn{1}{r}{\textit{Gold Answer}} & 2018 \\
    \multicolumn{1}{r}{Qwen3-4B} & \cellcolor{gray!15} The information provided in the documents does not specify when Season 3 of The Expanse started on Netflix. However, Season 3 of The Expanse premiered on Syfy in \uline{2018}, and Seasons 1 and 2 were streamed on Netflix until September 30, 2018. This implies that Season 3 may have been available on Netflix after September 30, 2018, but there is no explicit start date. \color{red}{(mentions the answer but does not answer)} \\
    \multicolumn{1}{r}{\method{}} & Season 3 of The Expanse started streaming on Netflix on September 7, 2018. \\
    \bottomrule
    \end{tabular}
\end{table*}

\paragraph{Failure Cases of Span Matching.}
ACC can also overestimate performance: a model may mention the correct answer somewhere in a long response while ultimately giving an incorrect, ambiguous, or self-contradictory final answer. 
As shown in Table~\ref{tab:llm_judge_cases}, such outputs can be counted as correct by lexical ACC because they contain the gold span, even though they reject, contradict, or over-qualify the answer. 
Therefore, we additionally introduce LLM-as-a-Judge to assess whether the prediction is truly semantically correct with respect to the question and reference answer. 
This helps filter out cases where the model merely guesses or incidentally includes the gold answer, providing a more reliable estimate of answer correctness.

\section*{Ethical Considerations}

Our work does not introduce new datasets containing personally identifiable or sensitive information, and we rely primarily on publicly available benchmarks. Nevertheless, context compression may introduce a risk of information loss, particularly when the compressed representation omits details that are important for downstream decision-making. This risk may be more consequential in sensitive or high-stakes applications, where seemingly minor information loss could affect the reliability of model outputs. Therefore, our method should be used with appropriate safeguards and human oversight in such settings. We also note that context compression does not provide any additional privacy guarantees, and sensitive information present in the original context may still be preserved in the compressed representation.

\bibliographystyle{ACM-Reference-Format}
\bibliography{latex/custom}

\end{document}